\documentclass[journal]{IEEEtran}
\usepackage[table,xcdraw]{xcolor}
\usepackage{graphicx}
\usepackage{listings}
\usepackage{bm}
\usepackage{pgfplots}
\usepackage{tikz}
\pgfplotsset{compat=newest}
\usepgfplotslibrary{groupplots}
\usepackage[utf8]{inputenc}
\usepackage{algpseudocode}
\usepackage{algorithm}
\usepackage{amsmath}
\usepackage{accents}
\newcommand{\norm}[1]{\left\lVert#1\right\rVert}
\usepackage{booktabs}
\begin {document}
\title {Classification Using Link Prediction}
\author{Seyed Amin Fadaee, Maryam Amir Haeri\\
{ \normalsize
\textit{Department of Computer Science and Information Technology,}\\
\textit{Amirkabir University of Technology,}\\
\textit{aminfadaee@aut.ac.ir, haeri@aut.ac.ir}\\
}
}
\maketitle

\begin{abstract}
Link prediction in a graph is the problem of detecting the missing links that would be formed in the near future. Using a graph representation of the data, we can convert the problem of classification to the problem of link prediction which aims at finding the missing links between the unlabeled data (unlabeled nodes) and their classes. To our knowledge, despite the fact that numerous algorithms use the graph representation of the data for classification, none are using link prediction as the heart of their classifying procedure. In this work, we propose a novel algorithm called CULP (\emph{C}lassification \emph{U}sing \emph{L}ink \emph{P}rediction) which uses a new structure namely Label Embedded Graph or LEG and a link predictor to find the class of the unlabeled data. Different link predictors along with Compatibility Score - a new link predictor we proposed that is designed specifically for our settings - has been used and showed promising results for classifying different datasets. This paper further improved CULP by designing an extension called CULM which uses a majority vote (hence the M in the acronym) procedure with weights proportional to the predictions' confidences to use the predictive power of multiple link predictors and also exploits the low level features of the data. Extensive experimental evaluations shows that both CULP and CULM are highly accurate and competitive with the cutting edge graph classifiers and general classifiers.
\end{abstract}


\section{Introduction}
Classification is an old problem in machine learning and pattern recognition that aims at finding a correct mapping between data and their corresponding labels. This mapping would then be used to derive the class of the unlabeled data \cite{murphy}.

This field is still highly active in the literature and a lot of algorithms have been proposed to correctly classify the data. Most of the classification algorithms aim at finding a decision boundary in the feature space for distinguishing the data belonging to different classes; however, as more complex data require more complex algorithms, these approaches could fail or not capture the true relations in the data.

One of the new approaches that has recently gained popularity in the literature is classification of the unlabeled instances using the graph representation of the data. Data can be represented in different forms one of which is a graph. In this setting, the data is first converted to a graph via a similarity function in the feature space, then unlabeled data is classified by incorporating a graph property. These graph properties are called \emph{high} level feature which give more insight to the data compared to the \emph{low} level features.

Classification using graph representation is studied extensively in numerous works (\cite{schemeDataClassificationUsingRandomWalk}, \cite{networkBasedHighLevelDataClassification}, \cite{lowHighLevelStacking}, \cite{highLevelTotallyBasedOnComplexetworks}, \cite{nonparametricClassificatioKAssociatedGraphs}, \cite{entropyHighLevel}, \cite{modularityHighLevel}, \cite{dataClassificationBasedOnImportance}). These works use graph properties such as clustering coefficient, modularity, importance, PageRank and others to classify the unlabeled data and they tend to achieve more accurate results compared to the classifiers that classify based on the low level features of data. This approach has been used in text classification \cite{text1}, hyperspectral image classification \cite{hyper1}, \cite{hyper2}, image classification \cite{schemeDataClassificationUsingRandomWalk}, \cite{modularityHighLevel}, handwritten digits recognition \cite{networkBasedHighLevelDataClassification} and other areas.

Link prediction is the problem of predicting the missing link that would be formed in the graph in near future \cite{linkpredictionsurvey}. Using the graph representation of the data we can treat the classification as a link prediction problem in an intuitive way where we try to find the link between the unlabeled node with it corresponding class. To our knowledge, there are not any work in the literature that uses link prediction to solve the problem of classification, however, the use of classification to solve link prediction is studied extensively \cite{linkpredictionsurvey}.

In this work, we proposed an algorithm called \emph{CULP} (acronym for Classification Using Link Prediction) that takes a different look at the classification problem through a link prediction approach. As we will elaborate in the paper, CULP uses a graph called LEG that models the data in an intuitive and suitable way for link prediction. 

Any link predictors can be used to derive the class of the unlabeled node in CULP and we proposed a new local measure called \emph{Compatibility Score} that is designed to improve the accuracy of link prediction and consequently classification.

As much insight as high level features have for capturing the patterns present in the data, exploiting the low level feature alongside them would further improve the predictive power of a graph classifiers and different researchers incorporate this idea in their work (\cite{schemeDataClassificationUsingRandomWalk}, \cite{lowHighLevelStacking}). This is why we further improved CULP and proposed the \emph{CULM} extension - a majority vote system (hence the \emph{M} in the acronym) with weights proportional to the probabilities of the predictions, this extension uses multiple link predictors along with a low level classifier. As we will see both CULP and CULM algorithms derive highly accurate results which are competitive with low level classifiers and other graph based classification methods.

The rest of the paper is organized as follows; in the next section a review of the general domains used in this paper is presented which is a preliminary section elaborating the problem of link prediction, similarity measures in vector space, method of converting graph to data and the problem of classification. After that a section of related works is given which is a summary of recent works using graph representation of the data for classification. Next, the CULP algorithm is presented with full details which elaborates on the \emph{LEG} (Label Embedded Graph) structure, the classification procedure which uses link prediction, our novel link predictor - Compatibility Score, the time complexity and a toy example to demonstrate CULP. Finally, the CULM extension is presented which is followed by our extensive experimental results to put our proposed algorithms into perspective. At the end, the conclusion to the paper and the aim for future works are presented.

\section{Preliminaries}
To fully understand CULP, a grounding for the details comprising this algorithm should be set. In this section, a general review to graph theory concepts and notations along with the definition of the link prediction problem in complex networks is given. After that, an overview of some of the most important similarity measures is presented, following this the different ways of converting data to graph is discussed. Finally at the end of this section the problem of classification is defined.
 
\subsection{Link Prediction}
Given a set of vertices $V$ and a set of edges $E$ containing $(i,j)$ where $i,j\in V$ the data structure $G(V,E)$ can be defined as a graph. If the elements in $E$ are ordered pairs, $G$ is considered to be a \emph{directed graph}. In an \emph{undirected graph} if $(i,j)\in E$ it is implied that $(j,i)\in E$. Regardless of the directionality of the graph, node $j$ is a \emph{neighbor node} to node $i$ if $(i,j)\in E$. For a node $i$, $\Gamma_i$ is the set of the neighbor nodes of $i$.

For the graph $G$, adjacency matrix $A_G$ or simply $A$ is defined as an $N\times N$ matrix with zero-one elements and $N=|V|$. For any entry in $A$, $A_{i,j}=1$ if and only if $(i,j)\in E$. In an undirected graph by definition $A=A^T$. As our focus in this paper is toward undirected graph, for the sake of simplicity we use \emph{graph} to state an undirected graph.

The degree of a node $i$ in a graph can be derived using $|\Gamma_i|$. For any graph, the cardinality or $|E|$ can be obtained by summing over the degree of all nodes using Equation \ref{cardinality} where $N=|V|$.
\begin{equation}\label{cardinality}
|E|=\frac{1}{2}\sum_{i=1}^{N}|\Gamma_i|
\end{equation}

The problem of link prediction in a graph arises when the goal is to predict for the currently absent links ($0$ entries in $A$) the probability of link formation in the future. There are many functions to predict the link prediction scores. These functions usually compute the local similarity between the nodes to derive the scores. One of the simplest techniques is known as \emph{common neighbors} (CN) \cite{commonNeighbors}. Using this approach the prediction scores can be derived using the following:

\begin{equation}\label{eq:CN}
\lambda_{i,j} =|\Gamma_i \cap \Gamma_j|
\end{equation}

Equation \ref{eq:CN} simply counts the number of common neighbors of nodes $i$ and $j$ to derive a score for their link formation. 

Another approach to find the link formation score is introduced by Adam and Adar \cite{adamicAdar} which uses degrees of common neighbors as features for prediction and it can be written as

\begin{equation}\label{eq:AA}
\lambda_{i,j}=\sum_{\gamma\in \Gamma_i\cap\Gamma_j}\frac{1}{log|\Gamma_\gamma|}
\end{equation}

Equation \ref{eq:AA} is known as the \emph{Adamic-Adar score} (AA). This score penalizes the features by their logarithm and uses these features for deriving the prediction scores. Another famous approach for tackling the problem of link prediction is the \emph{Resource Allocation Index} (RA) \cite{resourceAllocation} that simulates the transition of resources between nodes $i$ and $j$. This index is defined as Equation \ref{eq:RA}.

\begin{equation}\label{eq:RA}
\lambda_{i,j}=\sum_{\gamma\in \Gamma_i\cap\Gamma_j}\frac{1}{|\Gamma_\gamma|}
\end{equation}

This index is quite similar to AA, however it does not use the logarithm function which reduces the effect of nodes with high degree. This has the benefit of penalizing high degree common nodes. In a lot of networks, these nodes provide little insight for link prediction as they are connected to a lot of other nodes in the graph. 

In this work, we are proposing a new similarity function used for the purpose of link prediction. called \emph{Compatibility Score} which is discussed further in the paper.

\subsection{Similarity Measures}
Any data point $x$ with numeric features $x_f$ where $1\leq f \leq d$ can be regarded as a vector in an $d$-dimensional space. This view would enable the measurement of the similarities between data points using conventional similarity measures. As we are going to utilize a similarity measure in converting our data to graph(discussed in the next segment), we are going to provide overview of some of these measures.

Having our data matrix $X$, with $n$ rows and $d$ columns with each row being a data vector, the \emph{Cosine} similarity can be defined as the following:

\begin{equation}\label{eq:cosine}
s_{i,j}=\frac{X_i.X_j}{\norm{X_i}_2\norm{X_j}_2}
\end{equation}

Where $\norm{x}_2$ denotes the Euclidean norm of the vector x which is derived by the following:

$$\norm{x}_2=\sqrt{\sum_{f=1}^{d}x_f^2}$$

Following the above equation, the Euclidean distance between any two $d$ dimensional vectors can be written as:

\begin{equation}\label{eq:eqDist}
\phi_{i,j}=\sqrt{\sum_{f=1}^{d}(X_{i,f}-X_{j,f})^2}
\end{equation}

Utilizing the Euclidean distance, another similarity measure - namely \emph{Inverse Euclidean} can be defined using:

\begin{equation}\label{euclidean}
s_{i,j}=\frac{1}{\phi_{i,j}+\epsilon}
\end{equation}

In Equation \ref{euclidean} the $\epsilon$ term is a small number used to avoid division by zero in case of identical vectors. Another prominent distance in linear algebra is what is known as the absolute or Manhattan distance (Equation \ref{eq:manDist}) and by substituting Equation \ref{eq:manDist} in Equation \ref{euclidean}, the \emph{Inverse Manhattan} similarity function is defined.

\begin{equation}\label{eq:manDist}
\phi_{i,j}=\sum_{f=1}^{d}|X_{i,f}-X_{j,f}|
\end{equation} 

\subsection{Converting Data to Graph}
Any vector based data can be represented as a graph. Doing this would result in changing the structure of the data which enables us to compute high level features.

Two of the most used procedures for converting data to graph are $r$-Radius and $k$NN methods \cite{graphConversion}.

\begin{algorithm}[!h]
\caption{Undirected $k$NN conversion function for the data matrix $X$ and similarity measure $s$}
\label{knnAlgorithm}
\begin{algorithmic}
\Function{kNN-Convert}{$X$, $s$, $k$}
    \State $E = \{\}$
    \For {$i,j\in \{1,2,...,N\}$}
        \If {$i\in kNN(s,j)$  \textbf{or} $j\in kNN(s,i) $}
            \State $E\gets E\cup(i,j)$
        \EndIf
    \EndFor
    \State \Return $E$
\EndFunction
\end{algorithmic}
\end{algorithm}

Using a similarity measure (e.g. cosine similarity discussed in the previous segment) $s$ and matrix data $X$ we can use either of these two algorithms to convert the data into a graph. In $r$-Radius, an edge is created between every pair of data points that have a similarity higher than a predefined threshold $r$. Another approach is using $k$-nearest neighbors to form up the graph. If (based on a $s$) $X_i$ is in the $k$-nearest neighbors of $X_j$ the edge $(i, j)$ is created.

Due to the fact that $k$NN relation is not symmetric this approach would generally results in a directed graph. However the same principle can be used to create an undirected graph as in Algorithm \ref{knnAlgorithm}. Using this approach, if $X$ has $N$ instances, the number of undirected edges $|E|$ in the created graph is bounded by $\frac{Nk}{2} \le |E| \le Nk$. CULP uses an undirected $k$NN modeling of the data for the task of classification.

\subsection{Classification}
Suppose there are two sets of data, $X$ with $n$ instances and $d$ features for each instance which is the set of our labeled data. The labels of $X$ is denoted by $y$  where $y_i\in {1,2,...,C}$ with $C$ being the number of classes. Each pair $(X_i, y_i)$ makes up our training data. The other set of data is $X^{(u)}$ with $m$ instances and again $d$ features for each instance which are the unlabeled or the test data. 

The classification problem aims at finding a mapping $X^{(u)}_i \rightarrow \hat{y}_i$ for every $i\in{1,...,m}$. In other words, we are trying to find a proper label for each of the unlabeled instance in  $X^{(u)}$. If $C=2$, this is called \emph{binary classification} and if $C>2$, the problem is called \emph{multi-class classification} \cite{murphy}.

Classifiers like $k$NN or \emph{Decision Tree} can naturally handle multi-class classification problems, however some classifiers like \emph{SVM} are inherently designed for the binary classification task and upgrading them to handle multi-class classification requires using \emph{One vs. All} or \emph{One vs. One} approaches \cite{murphy}. 

In one vs. all, $C$ classifiers are trained and each classifier has the task of deciding whether an instance belongs to a particular class or not. The one vs. one approach is done by training $C(C-1)/2$ classifiers to classify an instance into either of two classes among all of the $C$ classes.

\section{Related Works}
Using graph classification has recently gained popularity and numerous works (\cite{schemeDataClassificationUsingRandomWalk, networkBasedHighLevelDataClassification, lowHighLevelStacking, highLevelTotallyBasedOnComplexetworks, nonparametricClassificatioKAssociatedGraphs, entropyHighLevel, modularityHighLevel}) focus on using this approach instead of the classical methods of classification . These method can capture complex patterns in the data and they can generate high level features to guide the classification procedure, furthermore they can usually be modified to utilize the low level features of the data as well.

In \cite{schemeDataClassificationUsingRandomWalk} a random walker is used to classify unlabeled instances on the graph embedding of the data. This graph is represented by a weight matrix of similarities. The random walk process is continued until convergence and the new data receives the label through a weighted majority vote between the labels of the top $\eta$ nodes with highest probabilities. This method takes the smilarity among the data points into account with a single network for the dataset along with structural changes of an unlabeled instance on the networks created for each class. The complexity of the method is of $O(n^2)$, however, as the authors claimed, using sparse representations such as $k$NN network, and graph construction method based on Lanczos bisection \cite{lanczosBisection}, this complexity can be reduced to a complexity between $O(n^{1.06})$ and $O(n^{1.33})$.

Another system is proposed in \cite{dataClassificationBasedOnImportance} in which a graph is created for the training instances of each class, then using the proposed \emph{spatio-structural differential efficiency} measure in the paper, a test instance is connected to some of the nodes in each graph. The label of the data would be the class of the graph that the test data has the highest importance in. The importance is characterized by Google's PageRank measure of the network. The spatio-structural differential efficiency measure in  \cite{dataClassificationBasedOnImportance} takes considers both physical and topological properties of the data and the complexity of the proposed method is again of $O(n^2)$ which is once more reduced to a complexity between $O(n^{1.06})$ and $O(n^{1.33})$ by using graph construction method based on Lanczos bisection.

A hybrid method is proposed in \cite{networkBasedHighLevelDataClassification} that aids a typical classifier (such as \emph{$k$NN}, \emph{SVM} or \emph{Naive Bayes}) by using high level features. These high level features are the difference of some graph properties before and after inserting a new instance into the graph representation of the data of each class. The graph of each class is constructed using combination of $r$-radius and $k$NN graph conversion methods. The graph properties used in their work are \emph{assortativity}, \emph{network clustering coefficient} and \emph{average degree}. The label for the test instance is generated by a weighted combination of low level and high level features. The authors extended their work in  \cite{lowHighLevelStacking} by using two more high level features namely \emph{Normalized Average Distance among vertices} and \emph{coreness variability} and using a stacking procedure to learn the weight for each feature. Also \cite{highLevelTotallyBasedOnComplexetworks} extends the same work by discarding the use of any classical classifier and using a scheme that takes low level features techniques into account to filter irrelevant graphs of some of the classes.

Authors of \cite{nonparametricClassificatioKAssociatedGraphs} proposed a framework for classification using \emph{k-Associated Optimal Graph} for modeling the data and \emph{Bayes theorem} and computing a posterior probability for each class to classify new instances. Similar to $k$NN graph conversion method, k-Associated Optimal Graph computes the similarity of a data point with all of the training data, however, it would form an edge only if the points belong to the same class. This would result in having multiple component (and possibly more than one component for a class). The method furthermore tries to find a local $k$ for each class so that the resulting components get the maximal $Purity$ (a measure based on average degree of a component). This way the process of finding the parameter $k$ is conducted automatically which also make the complexity of the framework of $O(n^2)$. Another paper \cite{networkBasedHighLevelDataClassification} also uses the k-Associated graph in this paper along with the high level classification method of  \cite{networkBasedHighLevelDataClassification} to classify new instances.

Other methods using different graph measures have been produced as well. \cite{entropyHighLevel} uses \emph{dynamic entropy} for each weighted graph produced by $r$-radius where the weights denote the distance between data points. \cite{modularityHighLevel} utilizes the \emph{modularity} measure for classifying new instance that belongs to a pattern set of the same object in the training data. The label is derived by creating a $k$NN graph for each pattern set and choosing the label of the graph with lowest modularity change after insertion of the new data. Both of the methods in \cite{entropyHighLevel} and \cite{modularityHighLevel} have the complexity of $O(n^2)$.

The graph based classification methods in the literature mostly have three characteristics in common. Firstly they create a different graph for each classes of the data; this approach avoids finding meaningful pattern that may form by the similarities between points in different classes.

The second aspect these algorithms have in common is that they treat test instances individually and add them to the graph of each class and measure a graph property before and after the insertion. This makes the prediction of a new instance inefficient in presence of large amount of test data.

Lastly, the properties that these algorithms use for finding the differences before and after the insertion of the unlabeled data (e.g. clustering coefficient, average path etc.) are time consuming and their computation times are usually dependent on the graph size which can make them infeasible for large datasets.

Our proposed algorithm CULP and it's extension CULM solves the first and second issue by employing a novel graph representation called \emph{LEG} which treats classes as nodes along with training and test instances as a unified object and is discussed further in the paper. As for the third problem, since the label of a test instance is derived using link prediction measures (as discussed in the previous section), the classification of the unlabeled data is faster than the similar methods.

\section{CULP Algorithm}
\emph{CULP} (\textbf{C}lassification \textbf{U}sing \textbf{L}ink \textbf{P}rediction) is a classification method aimed to gain a higher accuracy in mulit-class classification task by exploiting the similarity among the data points. This algorithm employs the powers of graph representation and link prediction methods in complex networks to deal with this problem\footnote{The complete code of CULP in python can be found in \texttt{github.com/aminfadaee/culp}}. The overall structure of CULP is consisted of 2 stages:

\begin{enumerate}
  \item Creating the LEG structure $G$ from the data
  \item Classifying the test data using $G$
\end{enumerate}

In the first step we model our data into an augmented graph data structure called \emph{LEG} (\textbf{L}abel \textbf{E}mbedded \textbf{G}raph) which we call $G$. $G$ is a heterogeneous graph which incorporates the data, the classes and the similarity between them as a unified object. 

LEG essentially contains 3 sets of nodes and 2 set of links. The different type of nodes in $G$ are training nodes, testing nodes and class nodes, also a link between two data nodes denotes similarity between them and a link between a training node and a class node denotes the class membership of that node.

After creating $G$, we can convert the classification problem to the problem of predicting the class membership link of a testing node. By utilizing a link prediction algorithm in the next step, membership score for every testing-class pair of nodes is computed.

Each of the membership scores acts as a posterior probability. A label is chosen for a testing node based on these scores.

CULP procedure is depicted in Algorithm \ref{culp}. In the next segments each of the steps of the proposed algorithm is covered in more detail.

\subsection{LEG Representation}
The first step toward classification using CULP is creating the LEG representation. LEG is a heterogeneous graph with three sets of nodes:
\begin{itemize}
  \item Training nodes ($V_l$)
  \item Testing nodes ($V_u$)
  \item Class nodes ($V_c$)
\end{itemize}
and two sets of edges:
\begin{itemize}
  \item Similarity edges ($E_s$)
  \item Class membership edges ($E_c$)
\end{itemize}

Each set of nodes correspond to their analogous set of data i.e. $V_l$ contains $n$ nodes, $V_u$ contains $m$ nodes and $V_c$ contains $C$ nodes. 

The class membership edges are created based on the labeled data. $E_c$ contain edges $(i,j)$ where $i\in V_l$ and $j$ is the node representation of $y_i$, meaning that each training node is connected (without direction) to its corresponding class node. It should be noted that since the labels for the test data is not available, $E_c$ contains only pair of nodes from $V_l$ and $V_c$.

Unlike $E_c$, the members of $E_s$ are not obtained so trivially. $E_s$ is responsible for incorporating the similarities between instances of our data and the edges in this set are obtained by using a graph conversion algorithm. In this work the undirected version of $k$NN graph conversion (Algorithm \ref{knnAlgorithm}) is used. 

Edges in $E_s$ primarily connect two nodes in $V_l$ or a node from $V_u$ to one in $V_l$. However, there is no constraint on having an edge between two nodes in $V_u$, meaning that we can find the similarity between unlabeled data and connect them as well (as we have done in this work).

If the unlabeled data is not available at first or in case of a new unlabeled node $x^{(u)}$ this node is first added to the set $V_u$, after that the similarity edges between this node and other nodes of the graph is created through a linear similarity computation.

After creating all of the sets of nodes and edges, we can define the LEG $G(V,E)$ where $V=V_l \cup V_u \cup V_c$ and $E=E_s \cup E_c$. Although $G$ is inherently heterogeneous, we can treat it as a simple undirected graph. The procedure for creating $G$ is summarized in Algorithm \ref{LEG}. This algorithm takes the labeled and unlabeled data along with the parameter $k$ and the similarity measure $s$ and produces $G$ as the output.

\begin{algorithm}[!h]
\caption{LEG construction function for the data $X^{(l)}$, the labels $y$ and the unlabeled data $X^{(u)}$ with parameter $k$ and the similarity function $s$}
\label{LEG}
\begin{algorithmic}
\Function{LEG}{$X^{(l)}$, $X^{(u)}$, $y$, $s$, $k$}
    \State {$X = X^{(l)}\cup X^{(u)}$ 
    \State $V_l \gets \{1,2,...,n\}$ \footnotesize{//Nodes are represented by numbers}}
    \State $V_u \gets \{n+1,n+2,..., n+m\}$
    \State $V_c \gets \{n+m+1,n+m+2,...,n+m+C\}$
    \State $E_c \gets \{\}$
    \For {$i\in\{1,2,...,n\}$}
            \State $E_c \gets E_c \cup(i,n+m+ y_i)$
    \EndFor
    \State $E_s \gets $kNN-CONVERT$(X,s,k)$
    \State  $V \gets V_l \cup V_u \cup V_c$
    \State  $E \gets E_s \cup E_c$
    \State \Return $G(V, E)$
    
\EndFunction
\end{algorithmic}
\end{algorithm}

There are always $n$ edges belonging to $E_c$. The number of edges in $E_s$ however, has an upper and lower bound. The minimum number of possible edges in $E_s$ is obtained when the $k$NN procedure of each pair of points in $X$ ($X^{(u)}\cup X^{(l)}$) is symmetric - meaning that $\forall i\forall j$ , $i\in kNN(j)\leftrightarrow j\in kNN(i)$. The maximum number of edges in $E_s$ on the other hand is obtained when the $k$NN procedure is \emph{not} symmetric for any pair of nodes in $X$. Using these, the bounds on the number of edges in a LEG can be derived as Equation \ref{LEGBound}.
\begin{equation}\label{LEGBound}
n+\frac{k}{2}(n+m)\leq |E|\leq n+k(n+m)
\end{equation}

By the bounds in Equation \ref{LEGBound}, it can be stated that $G$ gives us a new low memory cost representation of the data.  The memory for the original data is of $O(n\times d+m\times d +n)$ for $X^{(l)}$, $X^{(u)}$ and $y$, but since it is usually the case that $k<<d$ for high dimensional data, LEG saves a lot of memory compared to using the original data for the task of classification.

Another aspect of LEG is the fact that we are incorporating all of our labeled and unlabeled data and class labels in a unified structure that enables us to find the labels of the test data via simple and efficient graph properties, specifically link prediction methods which is covered in the next segment.
\subsection{Classification}
As stated before, in classification, the goal is to find a mapping $X^{(u)}_i \rightarrow \hat{y}_i$ for every $i\in{1,...,m}$. Using the LEG representation, this problem can be reformatted as finding $j^*$ for $\forall i\in V_U$ so that the probability of $(i,j^*)\in E_c$ is maximized.

The new formulation means that edges will be added to the set $E_c$ by predicting the most probable membership link for every test node. This can be easily done via link prediction methods discussed before.

Using a local similarity measure $\lambda$ for link prediction (e.g. Adamic-Adar index), this problem can be solved using the following:
  
\begin{equation}\label{eq1} 
\begin{cases} 
\forall i\in V_u,\;\; E_c\leftarrow E_c\cup (i,j^*)\\\\
j^*= \underset{j\in V_c}{argmax}(\lambda_{i,j})
\end{cases}
\end{equation}

Although more complex link prediction methods (random walk, average path length etc.) can be used to solve the problem, the local similarity measures are not only extremely fast and efficient to compute but they also derive competitively accurate results as it will be discuss in the experiments. The pseudocode of CULP is depicted in Algorithm \ref{culp}.

\begin{algorithm}[h]
\caption{CULP Algorithm}
\label{culp}
\begin{algorithmic}
\Function{CULP}{$X$, $X^{(u)}$, $y$, $s$, $k$, $\lambda$}
    \State $G \gets LEG(X, X^{(u)}, y, s, k)$
    \State $\hat{y}\gets \{\}$
    \For {$i \in V_u$}
        \State $j^*\gets \underset{j\in V_c}{argmax}(\lambda_{i,j})$\\
        \State $\hat{y}_i \gets j^*-(n+m)$
    \EndFor
    \State \Return $\hat{y}$
\EndFunction
\end{algorithmic}
\end{algorithm}

\subsection{Compatibility Score}
In this work a novel local score for link prediction is formed which is designed specifically for the task of classification. This new similarity function is called \emph{Compatibility Score} and like Adamic-Adar and Resource Allocation scores penalizes the common neighbors, however, this penalization is done differently.

\begin{figure}[ht]
\centering
\definecolor{ubuntu}{rgb}{0.2,0.2,0.2}
\definecolor{light}{rgb}{0.8,0.8,0.8}
\begin{tikzpicture}
\begin{groupplot}[group style={group size=1 by 2 }, height=4.2cm, width=\columnwidth,]
\nextgroupplot[
hide x axis,
hide y axis,
title={{\footnotesize LEG 1}},
xmin=-1, xmax=13,
ymin=-1.7, ymax=0.2,
]
\path [draw=black] (axis cs:0,0)--(axis cs:2,0);
\path [draw=black] (axis cs:2,0)--(axis cs:6,-0.5);
\path [draw=black] (axis cs:2,0)--(axis cs:6,-1);
\path [draw=black] (axis cs:2,0)--(axis cs:6,-1.5);
\path [draw=black] (axis cs:2,0)--(axis cs:11.5,0);
\path [draw=black] (axis cs:6,-0.5)--(axis cs:11.5,0);
\path [draw=black] (axis cs:6,-1)--(axis cs:11.5,0);
\path [draw=black] (axis cs:6,-1.5)--(axis cs:11.5,0);

\node at (axis cs:0,0)[  fill=light,  circle,  scale=0.85,  minimum size=0.5cm,  text=black,  draw=black]{ $i$};
\node at (axis cs:2,0)[  fill=ubuntu,  circle,  scale=0.85,  minimum size=0.5cm,  text=white]{ $\gamma$};
\node at (axis cs:6,-0.5)[fill=ubuntu,  circle,  scale=0.91,  minimum size=0.5cm,  text=white]{$a$ };
\node at (axis cs:6,-1)[  fill=ubuntu,  circle,  scale=0.83,  minimum size=0.5cm,  text=white]{$b$ };
\node at (axis cs:6,-1.5)[  fill=ubuntu,  circle,  scale=0.9,  minimum size=0.5cm,  text=white]{ $c$};
\node at (axis cs:11.5,0)[  fill=white,  circle,  scale=0.7,  minimum size=0.5cm,  text=black,  draw=black]{ $j_1$};
\node at (axis cs:11.5,-1)[  fill=white,  circle,  scale=0.7,  minimum size=0.5cm,  text=black,  draw=black]{ $j_2$};
\nextgroupplot[
hide x axis,
hide y axis,
title={{\footnotesize LEG 2}},
xmin=-1, xmax=13,
ymin=-1.7, ymax=0.2,
]
\path [draw=black] (axis cs:0,0)--(axis cs:2,0);
\path [draw=black] (axis cs:2,0)--(axis cs:6,-0.5);
\path [draw=black] (axis cs:2,0)--(axis cs:6,-1);
\path [draw=black] (axis cs:2,0)--(axis cs:6,-1.5);
\path [draw=black] (axis cs:2,0)--(axis cs:11.5,0);
\path [draw=black] (axis cs:6,-0.5)--(axis cs:11.5,-1);
\path [draw=black] (axis cs:6,-1)--(axis cs:11.5,-1);
\path [draw=black] (axis cs:6,-1.5)--(axis cs:11.5,-1);

\node at (axis cs:0,0)[  fill=light,  circle,  scale=0.85,  minimum size=0.5cm,  text=black,  draw=black,]{ $i$};
\node at (axis cs:2,0)[  fill=ubuntu,  circle,  scale=0.85,  minimum size=0.5cm,  text=white]{ $\gamma$};
\node at (axis cs:6,-0.5)[fill=ubuntu,  circle,  scale=0.91,  minimum size=0.5cm,  text=white]{$a$ };
\node at (axis cs:6,-1)[  fill=ubuntu,  circle,  scale=0.83,  minimum size=0.5cm,  text=white]{$b$ };
\node at (axis cs:6,-1.5)[  fill=ubuntu,  circle,  scale=0.9,  minimum size=0.5cm,  text=white]{ $c$};
\node at (axis cs:11.5,0)[  fill=white,  circle,  scale=0.7,  minimum size=0.5cm,  text=black,  draw=black]{ $j_1$};
\node at (axis cs:11.5,-1)[  fill=white,  circle,  scale=0.7,  minimum size=0.5cm,  text=black,  draw=black]{ $j_2$};
\end{groupplot}

\end{tikzpicture}
\caption{Using AA or RA for predicting the formation of $(i,j_1)$ in both LEG's would result in the same score, however node $\gamma$ in the first case is more valuable for the prediction.} \label{CSExample}
\end{figure}
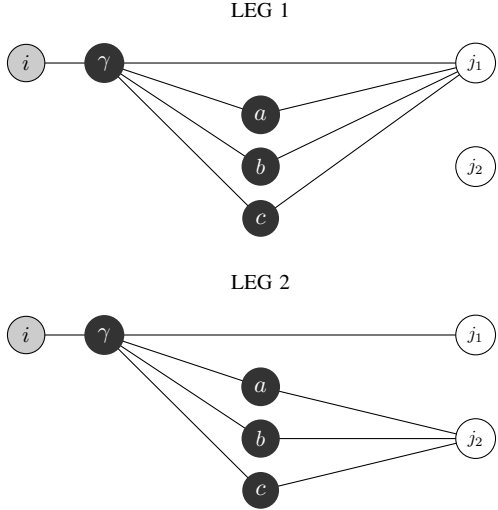

Both AA and RA scores can be unfair in some instances, meaning that they can over-penalize a valuable common neighbor or give the same score to two inherently different nodes. Take the two LEGs in Figure \ref{CSExample} for example ($i\in V_u$, $\gamma,a,b,c\in V_l$ and $j_1,j_2 \in V_c$). In both cases the goal is to find the score for the $(i,j_1)$ link. AA and RA would both penalize node $\gamma$ in the same way (penalty of $5$ for RA and $log(5)$ for AA); however, in the first LEG the node $\gamma$ is more valuable than that of the second LEG and this is due to the fact that three neighbors of this node ($a$, $b$, $c$) are also connected to node $j_1$.

When trying to predict the score for the formation of link between nodes $i$ and $j$ with a common neighbors between them namely $\gamma$, two sets of edges can be defined starting from $\gamma$: \emph{compatible edges} and \emph{incompatible edges}.

Compatible edges for node $\gamma$ are the ones connecting $\gamma$ to nodes which are by themselves connected to the destination of the candidate link ($j$ in this case). We can define incompatible edges as all the other edges which are not compatible.

Now the cardinality of incompatible edges or the \emph{incompatibility penalty} for node $\gamma$ which is a common neighbor of nodes $i$ and $j$ can be defined as the following:

\begin{equation}\label{eq:incom}
\delta(i,j,\gamma)=|\Gamma_\gamma|-|\Gamma_\gamma \cap \Gamma_j|
\end{equation}

Using Equation \ref{eq:incom} the Compatibility Score  (\emph{CS} for short) is formally defined as Equation \ref{eq:CS}. In this equation both $\delta(i,j,\gamma)$ and $\delta(j,i,\gamma)$ are used for the prediction of $(i,j)$ to make the score symmetric so that $\lambda_{i,j}=\lambda_{j,i}$.
\begin{equation}\label{eq:CS}
\lambda_{i,j}=\sum_{\gamma\in \Gamma_i\cap\Gamma_j}{\frac{1}{\delta(i,j,\gamma)}+\frac{1}{\delta(j,i,\gamma)}}
\end{equation}

Using the Compatibility Score for the cases of Figure \ref{CSExample} the score for link $(i,j_1)$ in LEG 1 can be computed as $0.7$ and in LEG 2 as $0.4$. This is the desired outcome as the score in LEG 1 is now higher. In the experiments, a more detailed comparison of CS with other link prediction methods is done.

\subsection{Time Complexity Analysis}
In this subsection, the time complexity of finding the class membership edge of a test node will be analyzed. The main component in finding the correct link is the local similarity measure $\lambda$ which is used for link prediction. These local measures find the score in time proportional to the degree of their source and destination nodes. In CULP, the source node $i$ belongs to $V_u$ and the destination node $j$ belongs to $V_c$. So the first step in analyzing the time of finding a class membership edge is finding the average degree of nodes in $V_u$ and $V_c$.

The degree of node $j$ is the number of labeled nodes connected to it or more specifically $n_j$ which is the number of data points with class of node $j$; however, for the degree of $i$ a more detailed analysis is needed. As stated before, in any undirected graph Equation \ref{cardinality} holds. This equation can be rewritten as the following:

$$|E|=\frac{1}{2}\left(\sum_{i\in V_c}{|\Gamma_i|}+\sum_{i\in V_l}{|\Gamma_i|}+\sum_{i\in V_u}{|\Gamma_i|}\right)$$

Since the degree of the class nodes sums up to the number of labeled data $n$, it can be substituted in the above equation; on the other hand, if we treat each node in $V_u$ to have average degree $D$, we can state that nodes in $V_l$ would have average degree of $D+1$ (since each of them has also a membership edge). Using all these, the above formula can be rewritten in the following manner:

$$|E|=\frac{1}{2}\left(n+n(D+1)+mD\right)$$
\begin{equation}\label{eq:start}
|E|=n+\frac{nD}{2}+\frac{mD}{2}
\end{equation}
As stated before the number of edges in a LEG is bounded by an upper and lower bound which is derived in Equation \ref{LEGBound}. Now using Equations \ref{eq:start} and \ref{LEGBound} the upper bound of $D$ can be defined as:
$$n+\frac{nD}{2}+\frac{mD}{2} = k(n+m)+n$$
\begin{equation}\label{eq:upper}
D=2k
\end{equation} 
and its lower bound as:
$$n+\frac{nD}{2}+\frac{mD}{2} =\frac{k}{2}(n+m)+n$$
\begin{equation}\label{eq:lower}
D=k
\end{equation} 
Consequently, the average degree for labeled and unlabeled nodes is of $O(k)$ and for class nodes is of $O(n)$. The \emph{Common Neighbor}, \emph{Adamic-Adar} and \emph{Resource Allocation} all have the complexity of finding the common neighbors between source and destination which is the intersection of the neighborhoods of the two nodes. The \emph{Compatibility Score} however, first finds the common neighbors and does two intersection for each of the nodes in the common neighbor set.

If done efficiently, the intersection of two sets with sizes $a$ and $b$ can be obtained in order of $O(min(a,b))$ in average. Using this, the complexity of finding the score in LEG for the formation of links between $i$ and $j$ is of $O(k)$ when \emph{Common Neighbor}, \emph{Adamic-Adar} or \emph{Resource Allocation} is used and is $O(k^2)$ when \emph{Compatibility Score} is used. Since $k$ is usually small (in our experiments $1\leq k\leq 35$), it is safe to state that the link prediction is done in constant time; also as there are $C$ nodes in $V_c$, predicting the label of $m$ instances would take time of $O(mC)$ after creating the LEG.
\subsection{Toy Example}
\begin{figure}[h]
\centering
\input{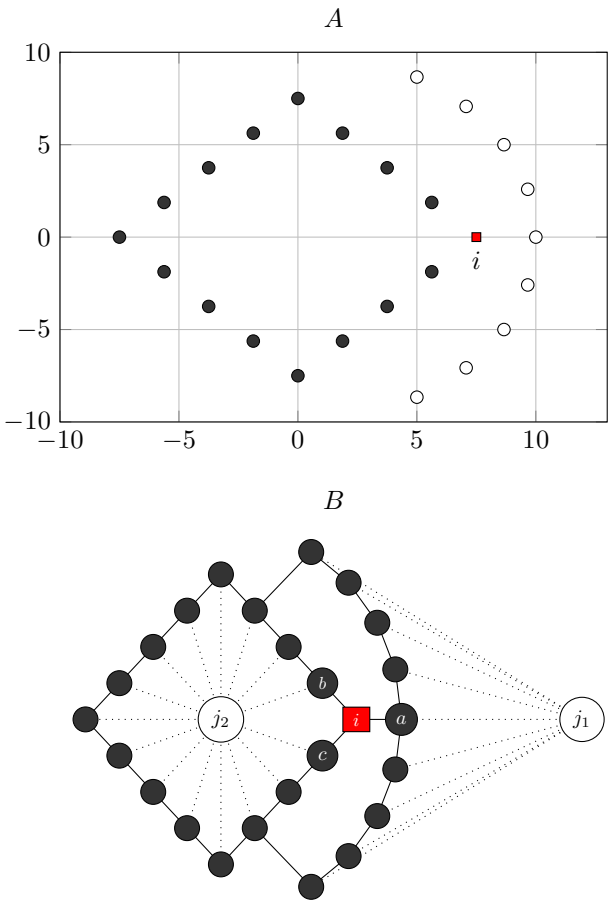}
\caption{Toy example demonstrating CULP. A- The set of data belonging to 2 classes and a test point in red B- LEG graph of the data} \label{toy}
\end{figure}

In this subsection a simple classification problem is solved using CULP to demonstrate the steps involving in this algorithm. The data is presented in Figure \ref{toy}-A as two classes. The white points represent the data of class 1 and the dark points belong to class 2. The problem is finding the correct label of the red point (point $i$).

The first step is choosing a similarity function $s$ and a value for the parameter $k$ for forming the graph. Here we chose $k=2$ and the Euclidean similarity (discussed in the preliminaries section).

Now the node sets can be defined as $V_c={j_1,j_2}$, $V_u={i}$ and all the other points as the set $V_l$. By creating the edges in $E_c$ and $E_s$ as shown in Algorithm \ref{LEG} the LEG in Figure \ref{toy}-B can be derived. As can be seen, in this graph every node except for $i$ is connected to one of the class nodes $j_1$ and $j_2$ (white nodes) by dotted links and the black links represents the edges of $E_s$.

Looking at the graph, it can be seen that the node $i$ is connected to nodes $a$, $b$ and $c$. This means these nodes would assist in finding the label for node $i$. Using these nodes, the scores for edges $(i,j_1)$ and $(i,j_2)$ can be obtained with each of the scores discussed before as $\lambda$. The results of computing these scores are depicted in Table \ref{toyTable}.

\begin{table}[ht]
\caption {Scores computed by 4 different link predictor for the toy example of Figure \ref{toy}} \label{toyTable}
\centering
\begin{tabular}{cllc}
\toprule
$\bm\lambda$    & $\bm{(i,j_1)}$              & $\bm{(i,j_2)}$ & \textbf{Prediction}\\
\midrule
\textbf{CN} & $1$&$2$&2 \\ 
\textbf{AA} &$1/log(4)$ &$2/log(3)$&2\\ 
\textbf{RA} &$1/4$ &$2/3$&2\\ 
\textbf{CS} &$1/2+1/4$&$2(1/2+1/3)$&2\\
\bottomrule      
\end{tabular}
\end{table}

The results of all the link predictors in Table \ref{toyTable} show that the score for the link $(i,j_2)$ is higher. This prediction matches the pattern perceived by looking at the data in Figure \ref{toy}-A and is the correct prediction.

\section{CULM Extension}
As we stated in the time complexity analysis subsection and demonstrated in the toy example of the previous section, once the LEG structure is formed, the prediction of links can be done instantly; knowing this and the fact that there are different options in choosing the link predictor $\lambda$, the question arises as to \emph{why not use all of our predictors and somehow combine their predictive capabilities to assist us in finding the best membership link for a test node?}

The next question arises after we analyze the related works done in the field of classification using complex network representations. A good portion of these methods are capable of incorporating or exploiting the low level features of the data to enhance the classification performance. \emph{How can we modify our framework CULP to exploit the low level features of the data as well as the high level features?}

The answer to both of these questions lies in our extension to CULP algorithm which we call the \emph{CULM} extension. CULM increases the predictive capabilities of CULP by using a \emph{weighted majority vote} procedure (hence the \textbf{M} as in \textbf{M}ajority in the end instead of \textbf{P}).

Instead of using only one link predictor $\lambda$, we will use an array of link predictors $\Lambda$. Each link predictor $\lambda$ when used, gives a score to the links $(i,j)$ for all $j\in V_c$. We can use all of these scores to estimate the probability $p$ of our prediction correctness as Equation \ref{confidence}.

\begin{equation}\label{confidence}
p_{\hat{y}}=\frac{\lambda_{i,j^*}}{\sum_{j\in V_c}{\lambda_{i,j}}}
\end{equation}

In this equation $\hat{y}$ is the label corresponding to $j^*$ and $j^*$ is computed using Equation \ref{eq1} of the previous section. Using Equation \ref{confidence} we can assign confidence to the prediction of $\lambda$. When using multiple predictors, it is obvious that a $\lambda$ with higher confidence is more reliable. We are going to use these probabilities to assign weights to each of the $\lambda$s  in $\Lambda$. This way instead of using a simple majority vote, a weighted voting procedure can be used. In a weighted majority vote procedure, few predictions are aggregated. Each of these prediction has an individual weight which states the value of their vote; finally the voting in this setting would be done as Algorithm \ref{voting}.

\begin{algorithm}[ht]
\caption{Weighted Majority Voting Algorithm}
\label{voting}
\begin{algorithmic}
\Function{VOTE}{$Y$, $W$}
    \State $L\gets \{0\}_C$
    \For {$y \in Y$ and $w\in W$ }
    		\State $L_y\gets L_y+w$
	\EndFor
	\State $\hat{y}\gets argmax(L)$
    \State \Return $\hat{y}$
\EndFunction
\end{algorithmic}
\end{algorithm}

In Algorithm \ref{voting}, $Y$ is the set containing the predicted labels of each of the predictors, $W$ is the respective weights of the labels and $L$ is a set with $C$ elements which keeps track of the weight for each of the classes. Using this algorithm enables us to not only use multiple link predictors' predicted labels, but also incorporate arbitrary any classical classifier $\phi$ with suitable weights. This way the low level features of the data is exploited as well.

\begin{algorithm}[ht]
\caption{CULM Algorithm}
\label{culm}
\begin{algorithmic}
\Function{CULM}{$X$, $X^{(u)}$, $y$, $s$, $k$, $\Lambda$, $\psi$, $\alpha$}
    \State $G \gets LEG(X, X^{(u)}, y, s, k)$
    \State $\hat{y}\gets \{\}$
    \For {$i$ in $V_u$}
    		\State $P\gets \{\}$
    		\State $\hat{Y}\gets \{\psi(X^{(u)}_i)\}$
    		\State $W\gets \{1-\alpha\}$
    		\For {$\lambda$ in $\Lambda$}
        		\State $j^*\gets \underset{j\in V_c}{argmax}(\lambda_{i,j})$\\
        		\State $P \gets P\cup \frac{\lambda_{i,j*}}{\sum\limits_{j\in V_c}{\lambda_{i,j}}}$
        		\State $\hat{Y} \gets \hat{Y}  \cup j^*-(n+m)$
    		\EndFor
    		\For {$p\in P$}
    			\State $W\gets W\cup \frac{\alpha\times p}{\sum\limits_{p'\in P}{p'}}$
    		\EndFor
    		\State $\hat{y}_i \gets \textrm{\textit{VOTE}}(\hat{Y},W)$
    \EndFor
    \State \Return $\hat{y}$
\EndFunction
\end{algorithmic}
\end{algorithm}

The next step is to define the weights for each of our predictors and $\phi$. If $\hat{y}_\lambda$ is the predicted label of the predictor $\lambda$ for the unlabeled data $x^{(u)}$ and $p_{\hat{y}}^\lambda$ is the probability of this prediction, the weight of predictor $\lambda$ for $x^{(u)}$ can be defined as Equation \ref{weight}. Also for the prediction of $\phi$ on $x^{(u)}$ which can be denoted as $\hat{y}_\phi$, we can define the weight as Equation \ref{weight_phi}.

\begin{equation}\label{weight}
w_{\hat{y}}^\lambda = \frac{\alpha p_{\hat{y}}^\lambda}{\sum\limits_{\lambda'\in\Lambda}{p_{\hat{y}}^{\lambda'}}}
\end{equation}

\begin{equation}\label{weight_phi}
w_{\hat{y}}^\phi = 1-\alpha
\end{equation}

The $\alpha$ parameter which is used in both equations is provided by the user. This parameter controls the trade-off that CULM will make between the link predictors' labels and the prediction of the low level classifier.

The parameter $\alpha$ is chosen in the range $0$ to $1$; however any value below $0.5$ would result in neutralizing the vote of CULM predictors. Also if $\alpha=1$, the prediction is completely done by CULM predictors and the low level classifier is ignored; so in general it can be stated that $0.5\leq \alpha \leq 1$.

Now the CULM extension can be formally defined as the procedure captured in Algorithm \ref{culm}. In this algorithm, after creating the LEG, each of the predictors in $\Lambda$ produce a label and a probability. These probabilities and labels are then merged with that of the low level classifier $\phi$ to form up $Y$ and $W$ which are passed to Algorithm \ref{voting} to produce the final label for the test instance. 

\section{Experimental Results}
In this section, we are presenting the result of our proposed algorithms CULP and CULM on 20 different real datasets and comparing it to classical classification methods as well as best classifiers of the related works in the domain of classification using complex networks.

The datasets used for our experiments are all obtained from UCI machine learning repository \cite{uci}. These datasets include \emph{Zoo}, \emph{Hayes-Roth} (Hayes), \emph{Iris}, \emph{Teaching Assistant Evaluation} (Teaching), \emph{Wine}, \emph{Sonar Mines vs. Rocks} (Sonar), \emph{Image Segmentation} training set (Image) and testing set (Segmentation), \emph{Glass Identification} (Glass), \emph{Thyroid Disease} (Thyroid), \emph{Ecoli}, \emph{Libras Movement} (Libras), \emph{Balance Scale} (Balance), \emph{Pima Indians Diabetes} (Pima), \emph{Statlog Vehicle Silhouettes} (Vehicle), \emph{Vowel Recognition} (Vowel), \emph{Yeast}, \emph{Wine Quality Red} (RedWine), \emph{Optical Recognition of Handwritten Digits} (Optical), \emph{Poker Hand} (Poker). Each of these datasets along with the number of instances, attributes and classes is listed in Table \ref{datasets}.

\begin{table}[h]
\caption{Datasets used in deriving the results for CULP and CULM.}\label{datasets}
\centering
\begin{tabular}{lccc}

\toprule
\textbf{Dataset}   & \textbf{Instances} & \textbf{Attributes} & \textbf{Classes}\\
\midrule
Zoo        & 101       & 16          & 7  \\
Hayes      & 132       & 4          & 3       \\
Iris       & 150       & 4          & 3       \\
Teaching   & 151       & 5          & 3       \\
Wine       & 178       & 13         & 3       \\
Sonar      & 208       & 60         & 2       \\
Image      & 210       & 19         & 7       \\
Glass      & 214       & 9          & 6       \\
Thyroid    & 215       & 5          & 3       \\
Ecoli      & 336       & 7          & 8       \\
Libras     & 360       & 90         & 15      \\
Balance    & 625       & 4          & 3       \\
Pima       & 768       & 8          & 2       \\
Vehicle    & 846       & 18         & 4       \\
Vowel      & 990       & 10         & 11      \\
Yeast      & 1,484      & 8          & 10      \\
RedWine & 1,599      & 11         & 6       \\
Segment    & 2,100      & 19         & 7       \\
Optical & 5,620      & 64         & 10      \\
Poker & 25,010     & 10         & 10    \\
  \bottomrule
\end{tabular}
\end{table}

\begin{table*}[t]
\caption{Results of CULP on the dataset with different link predictors. The number in parentheses represent the $k$ used in runs.}\label{result-culp}
\centering
\begin{tabular}{p{2cm}p{3cm}p{3cm}p{3cm}l}
\toprule
\textbf{Dataset}    & \textbf{CN}              & \textbf{AA}              & \textbf{RA}              & \textbf{CS}          \\
\midrule
Zoo         & 95.567 $\pm$ 5.8 (2)  & 96.567 $\pm$ 5.3 (2)  & \textbf{96.833 $\pm$ 5.4 (2)}  & 96.767 $\pm$ 5.4 (2)  \\
\midrule
Hayes       & \textbf{73.949 $\pm$ 12.0 (1)} & 73.718 $\pm$ 12.1 (1) & 73.718 $\pm$ 12.1 (1) & 73.667 $\pm$ 12.1 (1) \\
\midrule
Iris        & 98.467 $\pm$ 3.0 (11) & 98.467 $\pm$ 3.0 (11) & \textbf{98.489 $\pm$ 3.0 (11)} & 98.378 $\pm$ 3.2 (11) \\
\midrule
Teaching    & \textbf{63.756 $\pm$ 11.3 (1)} & 63.356 $\pm$ 11.1 (1) & 63.356 $\pm$ 11.1 (1) & 63.622 $\pm$ 11.2 (1)\\
\midrule
Wine        & 98.549 $\pm$ 2.8 (12) & \textbf{98.745 $\pm$ 2.6 (12)} & 98.725 $\pm$ 2.7 (12) & 98.137 $\pm$ 3.2 (12) \\
\midrule
Sonar       & \textbf{87.467 $\pm$ 7.4 (2)}  & 87.250 $\pm$ 7.3 (3)  & 86.900 $\pm$ 7.3 (3)  & 87.100 $\pm$ 7.5 (3)  \\
\midrule
Image       & 88.333 $\pm$ 6.7 (3)  & \textbf{89.317 $\pm$ 6.3 (3)}  & 89.175 $\pm$ 6.3 (3)  & 89.063 $\pm$ 6.4 (3)  \\
\midrule
Glass       & 71.857 $\pm$ 9.1 (3)  & 73.540 $\pm$ 9.2 (3)  & 73.397 $\pm$ 9.3 (2)  &\textbf{74.048 $\pm$ 9.1 (2) } \\
\midrule
Thyroid     & \textbf{97.540 $\pm$ 3.1 (4)}  & 97.413 $\pm$ 3.2 (4)  & 97.413 $\pm$ 3.2 (4)  & 97.333 $\pm$ 3.3 (4)  \\
\midrule
Ecoli       & 86.798 $\pm$ 6.1 (9)  & 87.010 $\pm$ 6.0 (8)  & \textbf{87.141 $\pm$ 6.1 (9)}  & 87.030 $\pm$ 6.0 (8)  \\
\midrule
Libras      & 79.935 $\pm$ 6.5 (2)  & 82.472 $\pm$ 6.3 (2)  & 81.713 $\pm$ 6.4 (2)  & \textbf{82.750 $\pm$ 6.2 (2)}  \\
\midrule
Balance     & 93.753 $\pm$ 2.9 (6)  & 96.446 $\pm$ 2.2 (2)  & 96.446 $\pm$ 2.2 (2)  & \textbf{96.780 $\pm$ 2.2 (2)}  \\
\midrule
Pima        & 76.061 $\pm$ 4.5 (34) & 76.154 $\pm$ 4.4 (28) & 76.211 $\pm$ 4.4 (28) & \textbf{76.355 $\pm$ 4.3 (7)}  \\
\midrule
Vehicle     & \textbf{73.611 $\pm$ 4.4 (5)}  & 73.091 $\pm$ 4.7 (5)  & 73.198 $\pm$ 4.7 (5)  & 72.512 $\pm$ 4.7 (5)  \\
\midrule
Vowel       & 97.603 $\pm$ 1.6 (3)  & \textbf{98.242 $\pm$ 1.5 (2)}  & \textbf{98.242 $\pm$ 1.5 (2)}  & 97.886 $\pm$ 1.5 (2)  \\
\midrule
Yeast       & 59.682 $\pm$ 3.9 (22) & 59.971 $\pm$ 3.7 (20) & 60.032 $\pm$ 3.6 (20) & \textbf{60.365 $\pm$ 3.8 (22)} \\
\midrule
RedWine & 60.501 $\pm$ 3.9 (1)  & 60.166 $\pm$ 3.9 (2)  & 60.036 $\pm$ 3.9 (2)  & \textbf{60.574 $\pm$ 3.8 (2)}  \\
\midrule
Segment     & 96.333 $\pm$ 1.3 (3)  & \textbf{96.535 $\pm$ 1.2 (4)}  & 96.525 $\pm$ 1.2 (4)  & 96.281 $\pm$ 1.3 (4)  \\
\midrule
Optical  & 98.805 $\pm$ 0.4 (5)  & 98.905 $\pm$ 0.4 (5)  & \textbf{98.918 $\pm$ 0.4 (5)} & 98.851 $\pm$ 0.4 (4)  \\
\midrule
Poker  &    58.518 $\pm$ 0.9 (32)   & 58.604 $\pm$ 0.9 (32) &  \textbf{58.625 $\pm$ 0.9 (32)}  &    58.520 $\pm$ 0.9 (32)     \\
\bottomrule      
\end{tabular}
\end{table*}
\subsection{CULP Analysis}
The reason behind choosing these datasets is the variety of both structure and domain between them. The size of these data is between 101 to 25,010 which test the practicality of our algorithms on both small and large datasets; the number of attributes vary from 4 to 90 which test the proposed algorithms against both low and high dimensional datasets and finally there is a lot of variety in the number of classes in the datasets which ranges from 2 up to 10.

This section is organized as follows: first, the experiment on CULP and different predictors as $\lambda$ is presented, after that the CULM algorithms is analyzed with 3 different low level classifier, the following subsection will discuss the effects of $\alpha$ parameter, after that a comparison of CULP and CULM with classical classifiers will be demonstrated and finally CULP and CULM will be compared along all the classical approaches and the similar works around classification using complex networks.

As the first experiment, different link predictors are used in CULP to compare the performance of each one on the datasets. For this experiments the predictor $\lambda$ is one of the CN, AA, RA and CS which are respectively defined in Equations \ref{eq:CN}, \ref{eq:AA}, \ref{eq:RA}, \ref{eq:CS}.

For each $\lambda$ and each dataset, the parameter $k$ ($1\leq k \leq 35$), the vector similarity function $s$ and a preprocessing procedure on the data (none, normalization or principle component analysis) is tuned. This tuning is done via a 10-Fold Cross Validation procedure. After finding the best parameters, 30 runs of 10-Fold Cross Validation is done that amount to total of 300 runs. Table \ref{result-culp} captures the results obtained by these settings.

In each cell of Table \ref{result-culp}, the first number is the mean accuracy of the runs and the second number is the standard deviation of them. The number in the parentheses represent the best $k$ obtained for each cell and the bold cell are the best result obtained on a dataset.

As can be seen in Table \ref{result-culp}, the Compatibility Score achieved the best results among the predictors, this is due to the fact that CS exclusively got the highest accuracy on 6 datasets of Glass, Libras, Balance, Pima, Yeast and RedWine. In the second place is the Resource Allocation Index that obtained the top accuracy for Zoo, Iris, Ecoli, Optical and Poker exclusively and achieved an identical best accuracy with Adamic-Adar Score on the Vowel dataset. The third best predictor is the Common Neighbor with 5 datasets of Hayes, Teaching, Sonar, Thyroid and Vehicle on top and finally Adamic-Adar for Wine, Image and Segment and the shared best results with RA for Vowel.

Analyzing the $k$s in this experiments, we can see that for 10 datasets of Zoo, Hayes, Iris, Teaching, Wine, Image, Thyroid, Libras, Vehicle and Poker the best $k$ is identical for each predictor on a dataset; in Balance and Pima however; the $k$s are noticeably different with Common Neighbor having the highest $k$ in both of them. In the rest of the datasets the choice of $k$ among different predictors are at most different by 1 (for Yeast it is 2).

\begin{table*}[t]
\caption{Results of CULM on the dataset with different link predictors. The first column is the best results obtained using CULP on each of the datasets. The number in parentheses represent the $k$  and $\alpha$ used in runs.}\label{result-culm}
\centering
\begin{tabular}{llllll}
\toprule
\textbf{Dataset} & \textbf{CULP} & \textbf{CULM}-\textit{LDA} & \textbf{CULM}-\textit{CART} & \textbf{CULM}-\textit{SVM} & \textbf{Gain}\\
\midrule
Zoo				& 96.833 $\pm$ 5.4 (RA, 2) 							& 97.467 $\pm$ 5.3 (1, 0.6) 							& \textbf{97.500 $\pm$ 5.0 (1, 0.6)}		&97.000 $\pm$ 5.9 (1, 0.6)			 									&+0.667\\
\midrule
Hayes       	&73.949 $\pm$ 12.0 (CN, 1)		 					& 74.513 $\pm$ 11.6 (1, 0.7) 							& \textbf{76.949 $\pm$ 11.1 (1, 0.6)}						&76.487 $\pm$ 11.1 (1, 0.6)				&+3.000 \\
\midrule
Iris 				&\textbf{98.489 $\pm$ 3.0 (RA, 11)} & 98.467 $\pm$ 3.0 (11, 0.7) 							& 98.467 $\pm$ 3.0 (11, 0.7) 						& 98.467 $\pm$ 3.0 (11, 0.7) 										&-0.022\\
\midrule
Teaching    &63.756 $\pm$ 11.3 (CN, 1) 							& \textbf{65.667 $\pm$ 11.6 (1, 0.6)} 	& 64.200 $\pm$ 12.0 (1, 0.6) 						& 65.622 $\pm$ 11.7 (1, 0.6)										&+1.911\\
\midrule
Wine			& 98.745 $\pm$ 2.6 (AA, 12) 						& \textbf{98.843 $\pm$ 2.9 (12, 0.7)} 	& 98.706 $\pm$ 2.7 (12, 0.7) 						& 98.745 $\pm$ 2.6 (12, 0.7) 										& +0.098\\
\midrule
Sonar       	&87.467 $\pm$ 7.4 (CN, 2) 							& 87.233 $\pm$ 7.2 (2, 0.6) 							&87.050 $\pm$ 7.4 (3, 0.7)								&\textbf{87.817 $\pm$ 7.3 (2, 0.6)} 					&+0.350\\
\midrule
Image       	&89.317 $\pm$ 6.3 (AA, 3) 							&\textbf{90.349 $\pm$ 6.2 (3, 0.6)}								&90.333 $\pm$ 6.0 (3, 0.6)	&89.571 $\pm$ 6.3 (3, 0.7)											&+1.032\\
\midrule
Glass       	&74.048 $\pm$ 9.1 (CS, 2)								&74.095 $\pm$ 9.1 (2, 0.6)								&\textbf{74.952 $\pm$ 8.8 (2, 0.6)}	&74.365 $\pm$ 9.4 (2, 0.6)											&+0.904 \\
\midrule
Thyroid     	& \textbf{97.540 $\pm$ 3.1 (CN, 4)}	&\textbf{97.540 $\pm$ 3.1 (4, 0.6)}		&97.492 $\pm$ 3.1 (4, 0.6)							&\textbf{97.540 $\pm$ 3.1 (4, 0.6)}					&0 \\
\midrule
Ecoli       		&87.141 $\pm$ 6.1 (RA, 9) 							&87.475 $\pm$ 5.8 (8, 0.6)		& \textbf{87.495 $\pm$ 5.9 (8, 0.6)}						&87.293 $\pm$ 5.8 (9, 0.6)											&+0.354 \\
\midrule
Libras      	&82.750 $\pm$ 6.2 (CS, 2) 							&82.843 $\pm$ 6.0 (2, 0.6)								&82.370 $\pm$ 5.8 (2, 0.6)							&\textbf{82.944 $\pm$ 5.9 (1, 0.6)}					&+0.194 \\
\midrule
Balance     	&96.780 $\pm$ 2.2 (CS, 2)								&97.016 $\pm$ 2.0 (2, 0.6)								&96.694 $\pm$ 2.1 (2, 0.7)							&\textbf{97.946 $\pm$ 1.7 (2, 0.6)}					&+1.166  \\
\midrule
Pima        	&76.355 $\pm$ 4.3 (CS, 7) 							&\textbf{76.535 $\pm$ 4.5 (7, 0.6)}		&76.461 $\pm$ 4.5 (7, 0.6)							&76.373 $\pm$ 4.6 (7, 0.6)											&+0.180 \\
\midrule
Vehicle     	&73.611 $\pm$ 4.4 (CN, 5)				 	&\textbf{74.829 $\pm$ 4.6 (5, 0.6)}		&73.897 $\pm$ 4.5 (5, 0.6)							&74.167 $\pm$ 4.6 (5, 0.6)											&+1.218 \\
\midrule
Vowel       &98.242 $\pm$ 1.5 (AA, 2)						&98.461 $\pm$ 1.3 (2, 0.9)								&98.508 $\pm$ 1.4 (2, 0.9)							&\textbf{98.620 $\pm$ 1.3 (2, 0.8)}					&+0.378 \\
\midrule
Yeast       	&\textbf{60.365 $\pm$ 3.8 (CS, 22)} 					&60.360 $\pm$ 3.6 (20, 0.6)	&60.288 $\pm$ 3.7 (20, 0.6)						&60.113 $\pm$ 3.7 (20, 1)											&-0.005 \\
\midrule
RedWine &60.574 $\pm$ 3.8 (CS, 2) 		&64.170 $\pm$ 3.7 (1, 0.6)								&63.453 $\pm$ 3.7 (1, 0.6)							&\textbf{64.447 $\pm$ 3.6 (1, 0.6)}					&+3.873 \\
\midrule
Segment     &96.535 $\pm$ 1.2 (AA, 4)			&96.673 $\pm$ 1.3 (2, 0.6)								&\textbf{96.922 $\pm$ 1.2 (2, 0.6)}	&96.651 $\pm$ 1.3 (2, 0.6)											&+0.387 \\
\midrule
Optical & \textbf{98.918 $\pm$ 0.4 (RA, 5)} &98.905 $\pm$ 0.4 (4, 0.9)						&98.890 $\pm$ 0.4 (4, 0.9)							&98.890 $\pm$ 0.4 (4, 0.9)											&-0.013 \\
\midrule
Poker  &58.625 $\pm$ 0.9 (RA, 32)		&58.581 $\pm$ 0.9 (32, 1)								&58.695 $\pm$ 0.9 (32, 0.6)						&\textbf{58.760 $\pm$ 0.9 (32, 0.6)}				&+0.135 \\
\bottomrule      
\end{tabular}
\end{table*}

\subsection{CULM Analysis} 
As the next experiment, the CULM algorithm is run on each of the datasets. The parameter $\alpha$ is tuned over the set $\{0.6, 0.7,0.8,0.9,1\}$. All the values below $0.6$ for $\alpha$ is not used to keep the results and comparisons fair (as stated before, any value below $0.5$ for $\alpha$ zeros the effect of CULP predictors also experimentally the same holds for $\alpha=0.5$), this way we are sure that the link predictors is not completely overshadowed by the low level classifier. Other parameters of the algorithm and the tuning is done as before and again each cell is the result of 300 runs.

For a low level classifier to accompany the link predictors in CULM, three different algorithms have been chosen and used. These low level classifiers are \emph{LDA} (Linear Discriminant Analysis), \emph{CART} (Classification And Regression Trees) and multi-class \emph{SVM} (Support Vector Machine) with RBF kernel.

Table \ref{result-culm} captures the results of this experiments. The first column is the best results for each of the datasets using CULP (Table \ref{result-culp}); the next three columns are the results of CULM with respectively  LDA, CART and SVM as $\phi$ and in each of the cells in these column the numbers in parentheses represent the $k$  and $\alpha$ used in runs. The last column in this table represents the accuracy gain achieved by using CULM instead of CULP. Each of the numbers in this column is obtained by comparing the best result obtained by CULM with the best result obtained by CULP for each dataset.

Looking at Table \ref{result-culm} it is clear that in the Thyroid dataset, using CULM achieved no change in the accuracy and in the datasets Iris and Optical the accuracy deteriorates; however, taking into account the other 17 datasets, CULM almost achieved a completely higher result.

CULM with SVM as its low level classifier achieved the best results on 6 datasets of Sonar, Thyroid, Libras, Balance, Vowel, RedWine and Poker exclusively and shares the best result on Thyroid with CULM-LDA and CULP. As the next best classifiers we have both CULM-CART and CULM-LDA with exclusively 5 best accuracy each (Zoo, Hayes, Glass, Ecoli and Segment for CULM-CART and Teaching, Wine, Image, Pima and Vehicle for CULM-LDA).

Datasets Hayes and RedWine achieved the highest accuracy gain (more than $3\%$) using CULM which is a noticeable boost. In the next level are datasets Teaching, Image Balance and Vehicle with more than $1\%$ gain. In general, the collective amount of gain achieved using CULM is the average of $0.8\%$ through all datasets which is another proof that CULM achieves a better results than CULP.

As for the parameter $k$, more robustness can be observed among different CULM classifiers than variations of CULP. Except for the datasets Sonar, Ecoli and Libras, the choice of $k$ in all variations of CULM are identical, also in these three datasets this parameter is different by at most $1$ on each classifier.

\begin{figure*}[t]
\centering
\begin{tikzpicture}

\begin{groupplot}[group style={group size=2 by 3,vertical sep=1.8cm, horizontal sep=2.5cm},height=\paperheight/6,width=\paperwidth/2.5]
\nextgroupplot[
title={Zoo},
xmin=0.37, xmax=1.03,
ymin=0.8987, ymax=0.978633333333333,
tick align=outside,
tick pos=left,
ticklabel style = {font=\tiny},
x grid style={white!69.01960784313725!black},
y grid style={white!69.01960784313725!black},
grid=major,
xlabel={\footnotesize $\alpha$},
ylabel={\footnotesize Accuracy}
] 
\addplot [ultra thick, red!50.19607843137255!black, forget plot]
table {%
0.4 0.920333333333333
0.5 0.920333333333333
0.6 0.974666666666667
0.7 0.974666666666667
0.8 0.974666666666667
0.9 0.974666666666667
1 0.952333333333333
}; 
\addplot [ultra thick, white!20.784313725490197!black, forget plot]
table {%
0.4 0.948666666666667
0.5 0.948666666666667
0.6 0.975
0.7 0.975
0.8 0.975
0.9 0.975
1 0.952333333333333
};
\addplot [ultra thick, white!66.66666666666666!black, forget plot]
table {%
0.4 0.902333333333333
0.5 0.902333333333333
0.6 0.97
0.7 0.97
0.8 0.97
0.9 0.97
1 0.952333333333333
};
\nextgroupplot[
title={Hayes},
xmin=0.37, xmax=1.03,
ymin=0.517538461538462, ymax=0.85374358974359,
tick align=outside,
tick pos=left,
ticklabel style = {font=\tiny},
x grid style={white!69.01960784313725!black},
y grid style={white!69.01960784313725!black},
grid=major,
xlabel={\footnotesize $\alpha$},
ylabel={\footnotesize Accuracy}
]
\addplot [ultra thick, red!50.19607843137255!black, forget plot]
table {%
0.4 0.532820512820513
0.5 0.532820512820513
0.6 0.744615384615385
0.7 0.745128205128205
0.8 0.745128205128205
0.9 0.745128205128205
1 0.737179487179487
};
\addplot [ultra thick, white!20.784313725490197!black, forget plot]
table {%
0.4 0.816666666666667
0.5 0.816666666666667
0.6 0.76948717948718
0.7 0.764871794871795
0.8 0.764871794871795
0.9 0.764871794871795
1 0.737179487179487
};
\addplot [ultra thick, white!66.66666666666666!black, forget plot]
table {%
0.4 0.838461538461538
0.5 0.838461538461538
0.6 0.764871794871795
0.7 0.762820512820513
0.8 0.762820512820513
0.9 0.762820512820513
1 0.737179487179487
};
\nextgroupplot[
title={Iris},
xmin=0.37, xmax=1.03,
ymin=0.937533333333333, ymax=0.986911111111111,
tick align=outside,
tick pos=left,
ticklabel style = {font=\tiny},
x grid style={white!69.01960784313725!black},
y grid style={white!69.01960784313725!black},
grid=major,
xlabel={\footnotesize $\alpha$},
ylabel={\footnotesize Accuracy}
]
\addplot [ultra thick, red!50.19607843137255!black, forget plot]
table {%
0.4 0.98
0.5 0.98
0.6 0.984444444444445
0.7 0.984666666666667
0.8 0.984666666666667
0.9 0.984666666666667
1 0.984666666666667
};
\addplot [ultra thick, white!20.784313725490197!black, forget plot]
table {%
0.4 0.939777777777778
0.5 0.939777777777778
0.6 0.984444444444445
0.7 0.984666666666667
0.8 0.984666666666667
0.9 0.984666666666667
1 0.984666666666667
};
\addplot [ultra thick, white!66.66666666666666!black, forget plot]
table {%
0.4 0.975333333333333
0.5 0.975333333333333
0.6 0.984
0.7 0.984666666666667
0.8 0.984666666666667
0.9 0.984666666666667
1 0.984666666666667
};
\nextgroupplot[
title={Teaching},
xmin=0.37, xmax=1.03,
ymin=0.524133333333333, ymax=0.662977777777778,
tick align=outside,
tick pos=left,
ticklabel style = {font=\tiny},
x grid style={white!69.01960784313725!black},
y grid style={white!69.01960784313725!black},
grid=major,
xlabel={\footnotesize $\alpha$},
ylabel={\footnotesize Accuracy}
]
\addplot [ultra thick, red!50.19607843137255!black, forget plot]
table {%
0.4 0.530444444444444
0.5 0.530444444444444
0.6 0.656666666666667
0.7 0.646
0.8 0.646
0.9 0.646
1 0.631555555555555
};
\addplot [ultra thick, white!20.784313725490197!black, forget plot]
table {%
0.4 0.635777777777778
0.5 0.635777777777778
0.6 0.640222222222222
0.7 0.637333333333333
0.8 0.637333333333333
0.9 0.637333333333333
1 0.631555555555555
};
\addplot [ultra thick, white!66.66666666666666!black, forget plot]
table {%
0.4 0.543777777777778
0.5 0.543777777777778
0.6 0.656222222222222
0.7 0.647333333333333
0.8 0.647333333333333
0.9 0.647333333333333
1 0.631555555555555
};
\nextgroupplot[
title={Wine},
xmin=0.37, xmax=1.03,
ymin=0.899029411764706, ymax=0.99371568627451,
tick align=outside,
tick pos=left,
ticklabel style = {font=\tiny},
x grid style={white!69.01960784313725!black},
y grid style={white!69.01960784313725!black},
grid=major,
xlabel={\footnotesize $\alpha$},
ylabel={\footnotesize Accuracy}
]
\addplot [ultra thick, red!50.19607843137255!black, forget plot]
table {%
0.4 0.989411764705883
0.5 0.989411764705883
0.6 0.987843137254902
0.7 0.98843137254902
0.8 0.98843137254902
0.9 0.98843137254902
1 0.986274509803922
};
\addplot [ultra thick, white!20.784313725490197!black, forget plot]
table {%
0.4 0.903333333333333
0.5 0.903333333333333
0.6 0.986078431372549
0.7 0.987058823529412
0.8 0.987058823529412
0.9 0.987058823529412
1 0.986274509803922
};
\addplot [ultra thick, white!66.66666666666666!black, forget plot]
table {%
0.4 0.983921568627451
0.5 0.983921568627451
0.6 0.987450980392157
0.7 0.987450980392157
0.8 0.987450980392157
0.9 0.987450980392157
1 0.986274509803922
};
\nextgroupplot[
title={Sonar},
xmin=0.37, xmax=1.03,
ymin=0.699841666666667, ymax=0.886658333333333,
tick align=outside,
tick pos=left,
ticklabel style = {font=\tiny},
x grid style={white!69.01960784313725!black},
y grid style={white!69.01960784313725!black},
grid=major,
xlabel={\footnotesize $\alpha$},
ylabel={\footnotesize Accuracy}
]
\addplot [ultra thick, red!50.19607843137255!black, forget plot]
table {%
0.4 0.741166666666667
0.5 0.741166666666667
0.6 0.872333333333334
0.7 0.865
0.8 0.865333333333333
0.9 0.865333333333333
1 0.865166666666667
};
\addplot [ultra thick, white!20.784313725490197!black, forget plot]
table {%
0.4 0.708333333333333
0.5 0.708333333333333
0.6 0.866166666666667
0.7 0.8685
0.8 0.868166666666667
0.9 0.868166666666667
1 0.865166666666667
};
\addplot [ultra thick, white!66.66666666666666!black, forget plot]
table {%
0.4 0.837666666666667
0.5 0.837666666666667
0.6 0.878166666666667
0.7 0.867666666666667
0.8 0.867333333333334
0.9 0.867333333333334
1 0.865166666666667
};
\end{groupplot} 
\end{tikzpicture}
\caption{Results of experimenting different values of $\alpha$ on 6 datasets. Each chart depicts accuracy on y axis and alpha on the x axis. Red lines are demonstrating the accuracy of CULM-LDA, black lines are CULM-CART and the gray lines depict the results of CULM-SVM. } \label{alpha}
\end{figure*}
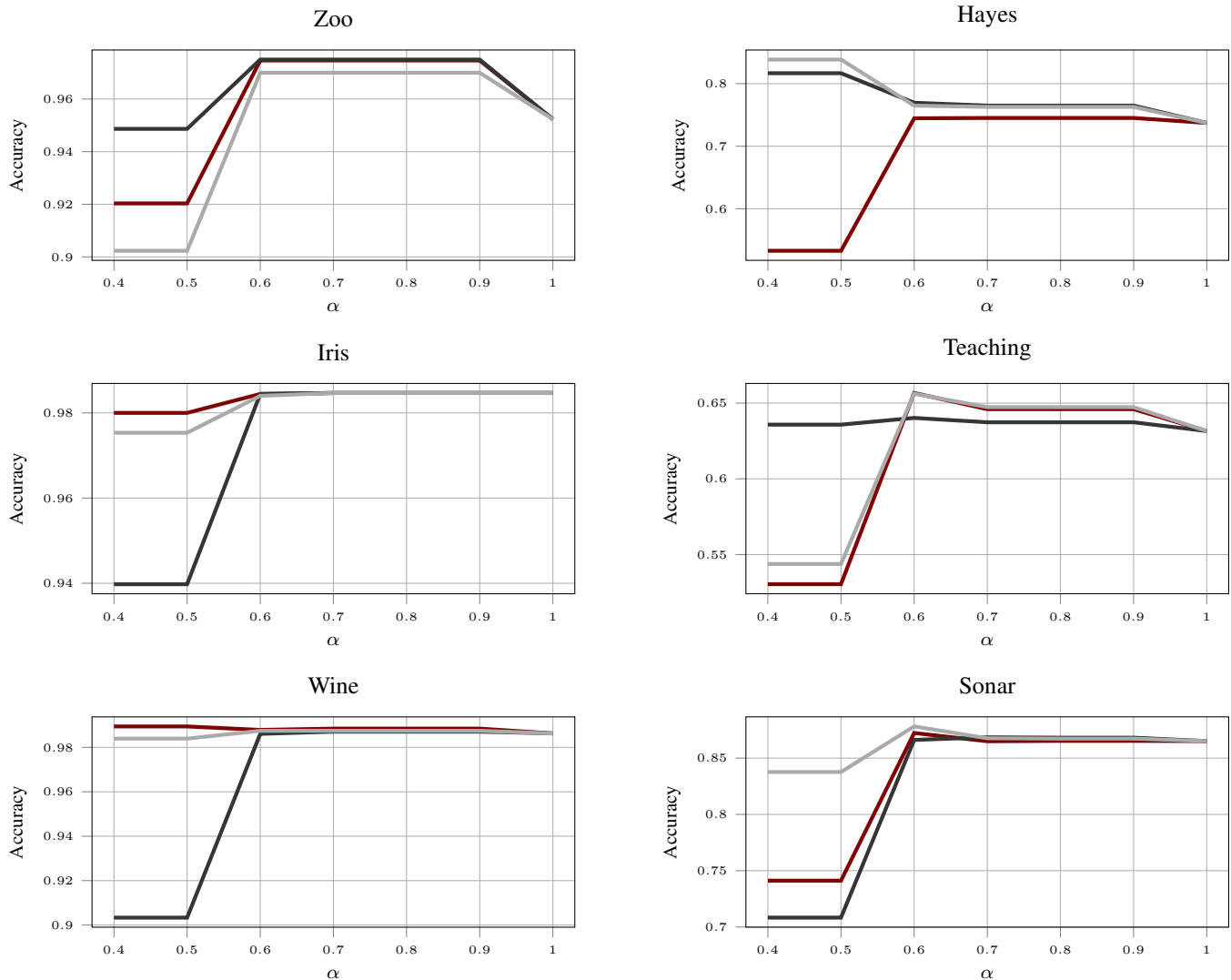 

The other parameter $\alpha$ in this experiments reveals interesting facts as well. Except for the CULM-SVM on Yeast data and CULM-LDA on Poker dataset, we can observe $\alpha<1$ in all the experiments; this shows that using the low level features through the low level classifier did indeed help the classification accuracy. Saying this, we still need a more detailed analysis on the effect of $\alpha$ on the accuracy which is the main discussion of the next segment.

\subsection{$\alpha$ Analysis}

To analyze the $\alpha$ parameter further, six datasets were chosen, each with a single configuration to run with different $\alpha$ values. The datasets are Zoo with $k=1$, Hayes with $k=1$, Iris with $k=11$, Teaching with $k=1$, Wine with $k=12$ and Sonar with $k=2$ and $\alpha\in\{0.4,0.5,0.6,0.7,0.8,0.9,1\}$  in each experiment. The choices for $\alpha$ is to demonstrate the effect of zeroing the effect of predictors ($\alpha\leq 0.5$), zeroing the effect of low level classifier ($\alpha =1$) or picking something in between.

The results of this experiment are depicted in the charts of Figure \ref{alpha}. Each chart represents the experiments done on a dataset. These charts capture the accuracy of each of the 3 CULM classifiers for each value of $\alpha$. Red lines are demonstrating the accuracy of CULM-LDA, black lines are CULM-CART and the gray lines depict the results of CULM-SVM.

As stated before, any value below $0.5$ for $\alpha$ zeros the effect of CULP predictors, we also noted that experimentally the same holds for $\alpha=0.5$. This is evident by looking at the plots of Figure \ref{alpha} because in all datasets and classifiers the accuracies obtained for $\alpha=0.4$ and $\alpha=0.5$ are identical.

As can be seen from the figure, for all classifiers of the datasets Zoo, Iris, Teaching and Sonar, using the predictors improved the accuracy of the low-level classifier; on the other hand, in all datasets zeroing the effect of the low-level classifier ($\alpha=1$) had not helped (if not worsened) the accuracy of the prediction. The other notable detail in these plot is the plateau of accuracy for roughly the values of $\alpha$ between $0.6$ and $0.9$. This means that a less fine-grained set of values can be also used for tuning this parameter.

\begin{table*}[t]
\caption{Results of comparing CULP and CULM with 4 different classical classifiers. The number in the parentheses in the cells of first three column represent $k$ and the bold cells are the best results obtained on the dataset.}\label{result-classic}
\centering
\begin{tabular}{lllllll}
\toprule
\textbf{Dataset} & \textbf{CULP} & \textbf{CULM} &\textbf{$k$NN}&\textbf{LDA}&\textbf{CART}&\textbf{SVM}\\
\midrule
Zoo				&96.833 $\pm$ 5.4 (2)                  &\textbf{97.500 $\pm$ 5.0 (1)}	           &97.500 $\pm$ 12.7 (1)       &92.033 $\pm$ 8.2     &94.867 $\pm$ 6.8     &90.233 $\pm$ 10.0\\
\midrule
Hayes       	&73.949 $\pm$ 12.0 (1)                  &76.949 $\pm$ 11.1 (1)           &72.590 $\pm$ 14.6 (1)      &53.282 $\pm$ 13.0     &81.667 $\pm$ 9.5    &\textbf{85.103 $\pm$ 8.4}\\
\midrule
Iris 			&\textbf{98.489 $\pm$ 3.0 (11)}               &98.467 $\pm$ 3.0 (11)             &97.222 $\pm$ 5.4 (17)     &98.000 $\pm$ 3.5        &94.556 $\pm$ 5.6    &97.533 $\pm$ 3.8\\
\midrule
Teaching        &63.756 $\pm$ 11.3 (1)                  &\textbf{65.667 $\pm$ 11.6 (1)}     &62.511 $\pm$ 12.7 (1)     &53.044 $\pm$ 13.2    &64.333 $\pm$ 11.4    &54.378 $\pm$ 12.7\\
\midrule
Wine			&98.745 $\pm$ 2.6 (12)                 &98.843 $\pm$ 2.9 (12)     &97.000 $\pm$ 5.6 (24)       &\textbf{98.941 $\pm$ 2.4}      &90.294 $\pm$ 6.9    &98.392 $\pm$ 3.0\\
\midrule
Sonar       	&87.467 $\pm$ 7.4 (2)                   &\textbf{87.817 $\pm$ 7.3 (2)}      &86.383 $\pm$ 8.0 (1)      &74.117 $\pm$ 9.1      &70.850 $\pm$ 9.7     &83.767 $\pm$ 7.8\\
\midrule
Image       	&89.317 $\pm$ 6.3 (3)                   &\textbf{90.349 $\pm$ 6.2 (3)}       &85.667 $\pm$ 8.3 (4)      &89.635 $\pm$ 6.1     &88.222 $\pm$ 6.8     &87.317 $\pm$ 6.4\\
\midrule
Glass       	&74.048 $\pm$ 9.1 (2)                   &\textbf{74.952 $\pm$ 8.8 (2)}      &72.683 $\pm$ 10.4 (1)     &62.381 $\pm$ 10.0     &66.698 $\pm$ 9.1    &70.190 $\pm$ 9.8\\
\midrule
Thyroid     	&\textbf{97.540 $\pm$ 3.1 (4)}	            &\textbf{97.540 $\pm$ 3.1 (4)}      &96.206 $\pm$ 5.8 (1)      &91.397 $\pm$ 5.7      &93.857 $\pm$ 5.4    &95.921 $\pm$ 4.0\\
\midrule
Ecoli       	&87.141 $\pm$ 6.1 (9)                   &\textbf{87.495 $\pm$ 5.9 (8)}      &86.909 $\pm$ 6.3 (7)      &86.869 $\pm$ 5.6      &79.515 $\pm$ 6.8    &86.828 $\pm$ 6.0\\
\midrule
Libras      	&82.750 $\pm$ 6.2 (2)                   &82.944 $\pm$ 5.9 (1)      &\textbf{85.880 $\pm$ 8.0 (1)}       &64.620 $\pm$ 8.4      &68.713 $\pm$ 8.2     &80.306 $\pm$ 6.6\\
\midrule
Balance     	&96.780 $\pm$ 2.2 (2)                   &\textbf{97.946 $\pm$ 1.7 (2)}      &90.140 $\pm$ 5.5 (15)      &86.747 $\pm$ 3.9     &81.306 $\pm$ 5.9     &90.489 $\pm$ 3.6\\
\midrule
Pima        	&76.355 $\pm$ 4.3 (7)                   &76.535 $\pm$ 4.5 (7)       &74.171 $\pm$ 4.8 (9)      &\textbf{77.320 $\pm$ 4.3}      &70.289 $\pm$ 5.1     &76.013 $\pm$ 4.4\\
\midrule
Vehicle     	&73.611 $\pm$ 4.4 (5)				 	&74.829 $\pm$ 4.6 (5)      &72.206 $\pm$ 5.2 (6)      &\textbf{78.052 $\pm$ 4.3}      &71.282 $\pm$ 4.9    &76.675 $\pm$ 4.8\\
\midrule
Vowel           &98.242 $\pm$ 1.5 (2)                   &98.620 $\pm$ 1.3 (2)      &\textbf{98.983 $\pm$ 2.3 (1)}      &59.556 $\pm$ 4.7      &81.192 $\pm$ 4.2    &94.852 $\pm$ 2.3\\
\midrule
Yeast       	&\textbf{60.365 $\pm$ 3.8 (20)} 					&60.360 $\pm$ 3.6 (20)     &59.586 $\pm$ 3.8 (19)     &58.923 $\pm$ 3.8     &51.205 $\pm$ 4.0     &60.124 $\pm$ 3.7\\
\midrule
RedWine     &60.574 $\pm$ 3.8 (2) 		        &64.447 $\pm$ 3.6 (1)      &\textbf{64.662 $\pm$ 3.8 (1)}      &59.172 $\pm$ 3.9     &63.390 $\pm$ 3.8      &62.637 $\pm$ 3.7\\
\midrule
Segment         &96.535 $\pm$ 1.2 (4)			    &\textbf{96.922 $\pm$ 1.2 (2)}      &95.829 $\pm$ 1.8 (1)      &91.446 $\pm$ 1.9     &95.459 $\pm$ 1.4     &93.825 $\pm$ 1.6\\
\midrule
Optical      &\textbf{98.918 $\pm$ 0.4 (5)}                &98.905 $\pm$ 0.4 (4)         &98.823 $\pm$ 0.5 (3)      &95.278 $\pm$ 0.8     &90.532 $\pm$ 1.3     &98.681 $\pm$ 0.5\\
\midrule
Poker      &58.625 $\pm$ 0.9 (32)		        &\textbf{58.760 $\pm$ 0.9 (32)}     &58.517 $\pm$ 1.0 (34)     &49.952 $\pm$ 0.9     &48.948 $\pm$ 1.7     &58.617 $\pm$ 0.5\\
\bottomrule      
\end{tabular}
\end{table*}
For the next experiment, the results of CULP and CULM is compared with 4 classical classifiers. These classifiers include $k$NN classifier, LDA, CART and multi-class SVM with RBF kernel.

\subsection{Comparison to Classical Classifiers}
The results of this experiment is captured in Table \ref{result-classic}. In this table the first column represent the best result of CULP for each dataset, the second column is the best result of CULM for each dataset and the other 4 columns are the results obtained by the classical classifier. The number in the parentheses in the cells of first three column represent $k$ and the bold cells are the best results obtained on the dataset.

Comparing the $k$ values in the first 3 columns of Table \ref{result-classic}, we can realize that except for Ecoli and Yeast, this parameter is smaller (or equal) for CULP and CULM than that of the $k$NN algorithm and in some cases like Wine and Balance this difference is quite high. This is due to the fact that the undirected version of $k$ nearest neighbor is used to form up the LEG graph which consequently enables us to capture the similarity features with less neighbors.

It is evident from the results that CULP and CULM achieved superior results compared to the classical algorithms. CULP and CULM collectively obtained the best results on 13 datasets. The $k$NN and LDA algorithms achieved the highest results on 3 datasets each, SVM got the best results only on the Hayes dataset and CART is completely outperformed by the other algorithms on all datasets.

One thing that can be noted is the fact that CULM could obtain the best results on the datasets Hayes, Wine, Pima and Vehicle with $\alpha\leq 0.5$ but as stated before we decided to forgo these values to give a fair comparison; however, in general we can state that CULM can outperform or achieve the same result of any classical classification algorithms given the right configuration for the $\alpha$ parameter.

\subsection{Complete Comparison}

\begin{table*}[t]
\caption{Results of comparing CULP and CULM, classical classifiers, HLCRW \cite{schemeDataClassificationUsingRandomWalk} and PgRkNN \cite{dataClassificationBasedOnImportance} algorithms.}\label{result-comp}
\centering
\begin{tabular}{p{2cm}p{2cm}p{2cm}p{2cm}p{2cm}p{2cm}}
\toprule
\textbf{Dataset} & \textbf{CULP} & \textbf{CULM}& \textbf{Classical} & \textbf{HLCRW}&\textbf{PgRkNN}\\
\midrule
Zoo				&96.833 $\pm$ 5.4                  &97.500 $\pm$ 5.0	           &97.500 $\pm$ 12.7      &   97.00 $\pm$ 0.1     &    \textbf{99.03 $\pm$ 2.9}      \\
\midrule
Hayes       	&73.949 $\pm$ 12.0                   &76.949 $\pm$ 11.1           &\textbf{85.103 $\pm$ 8.4}     &  61.70 $\pm$ 2.3 &   73.09 $\pm$ 11.7 \\
\midrule
Iris 			&\textbf{98.489 $\pm$ 3.0}               &98.467 $\pm$ 3.0            &98.000 $\pm$ 3.5       & 98.00 $\pm$ 0.6 &  97.20 $\pm$ 3.7  \\
\midrule
Teaching        &63.756 $\pm$ 11.3                  &\textbf{65.667 $\pm$ 11.6}     &64.333 $\pm$ 11.4    &  \underline{65.30 $\pm$ 2.0} &  62.08 $\pm$ 13.4 \\
\midrule
Wine			&98.745 $\pm$ 2.6                 &98.843 $\pm$ 2.9     &\textbf{98.941 $\pm$ 2.4}     &   87.10 $\pm$ 1.6  &    93.95 $\pm$ 5.3  \\
\midrule
Sonar       	&87.467 $\pm$ 7.4                   &\textbf{87.817 $\pm$ 7.3}      &86.383 $\pm$ 8.0     &  81.79 $\pm$ 7.8 &    82.00 $\pm$ 7.5  \\
\midrule
Image       	&89.317 $\pm$ 6.3                   &\textbf{90.349 $\pm$ 6.2}       &89.635 $\pm$ 6.1     &   75.60 $\pm$ 0.8&   86.13 $\pm$ 7.2  \\
\midrule
Glass       	&74.048 $\pm$ 9.1                   &\textbf{74.952 $\pm$ 8.8}      &72.683 $\pm$ 10.4    & 72.80 $\pm$ 1.1 &   71.75 $\pm$ 7.9   \\
\midrule
Thyroid     	&97.540 $\pm$ 3.1            &97.540 $\pm$ 3.1      &96.206 $\pm$ 5.8     &\textbf{97.57 $\pm$ 3.0}&    97.55 $\pm$ 3.0  \\
\midrule
Ecoli       	&87.141 $\pm$ 6.1                   &\textbf{87.495 $\pm$ 5.9}      &86.909 $\pm$ 6.3     &   85.50 $\pm$ 0.6 &    85.11 $\pm$ 5.4  \\
\midrule
Libras      	&82.750 $\pm$ 6.2                   &82.944 $\pm$ 5.9      &85.880 $\pm$ 8.0      &  85.00 $\pm$ 0.8 &  \textbf{87.16 $\pm$ 9.8} \\
\midrule
Balance     	&96.780 $\pm$ 2.2                   &\textbf{97.946 $\pm$ 1.7}      &90.489 $\pm$ 3.6     &  97.20 $\pm$ 0.6&   90.86 $\pm$ 3.4 \\
\midrule
Pima        	&76.355 $\pm$ 4.3                   &76.535 $\pm$ 4.5       &\textbf{77.320 $\pm$ 4.3}      &  75.54 $\pm$ 4.6 &   74.85 $\pm$ 4.9 \\
\midrule
Vehicle     	&73.611 $\pm$ 4.4			 	&74.829 $\pm$ 4.6      &\textbf{78.052 $\pm$ 4.3}     &  67.70 $\pm$ 0.6&    70.26 $\pm$ 4.1   \\
\midrule
Vowel           &98.242 $\pm$ 1.5                   &98.620 $\pm$ 1.3      &\textbf{98.983 $\pm$ 2.3}     &   97.50 $\pm$ 0.3 &  98.49 $\pm$ 1.2   \\
\midrule
Yeast       	&\textbf{60.365 $\pm$ 3.8}					&60.360 $\pm$ 3.6     &60.124 $\pm$ 3.7     &   57.20 $\pm$ 0.5  &   56.50 $\pm$ 3.6 \\
\midrule
RedWine     &60.574 $\pm$ 3.8 		        &64.447 $\pm$ 3.6      &64.662 $\pm$ 3.8     &  61.60 $\pm$ 0.5 &   \textbf{66.68 $\pm$ 3.5}\\
\midrule
Segment         &96.535 $\pm$ 1.2         &\textbf{96.922 $\pm$ 1.2}         &95.829 $\pm$ 1.8     & 93.20 $\pm$ 0.2 &   95.63 $\pm$ 1.5\\
\midrule
Optical      &98.918 $\pm$ 0.4                &98.905 $\pm$ 0.4         &98.823 $\pm$ 0.5     &    95.09 $\pm$ 2.1    &   \textbf{98.94 $\pm$ 0.4}  \\
\midrule
Poker      &58.625 $\pm$ 0.9		        &\textbf{58.760 $\pm$ 0.9}     &58.617 $\pm$ 0.5     &   55.42 $\pm$ 0.9     &      53.78 $\pm$ 0.8    \\
\bottomrule      
\end{tabular}
\end{table*}
As the final experiment of this paper, a complete comparison is done to analyze the results of CULM, CULP, the classical algorithms and two of the similar works that use complex network representation of the data to classify the unlabeled instances. These two classifiers which were discussed in the related work sections are PgRkNN \cite{dataClassificationBasedOnImportance} and HLCRW \cite{schemeDataClassificationUsingRandomWalk} (short for High Level data Classification using Random Walk)

The results of PgRkNN and HLCRW algorithms on datasets which were already provided in their papers are used here without a change, for other cases we implemented and run both of them completely by the details provided in those papers.

Table \ref{result-comp} captures these results along with the summaries of Tables \ref{result-culp}, \ref{result-culm} and \ref{result-classic}. For each of the rows in this table the bold cell is the best result for classifying the instances of the dataset through all of the algorithms. The best result for each of the cases where CULP/CULM obtained the higher average accuracy, is tested for significance against the second best accuracy using the Welch's \emph{t}-test with confidence level of 0.95. In this test, the null hypothesis is that the averages are the same and the alternative hypothesis is that they are different. Except for the Teaching dataset which the bold and underlined values are not significantly different all the other bold values in CULP and CULM columns are superior.

In the first glance at Table \ref{result-comp} it can be realized that CULM is the leader with 8 best results on the datasets among different algorithms. These datasets include Teaching, Sonar, Image, Glass, Ecoli, Balance, Segment and Poker.The next best algorithm in case of the best results is the Classical group with 5 dataset of Hayes, Wine, Pima, Vehicle and Vowel in lead. As the third algorithm we have PgRkNN with datasets Zoo, Libras, RedWine and Optical. The one before last is CULP with Iris and Yeast on top and finally HLCRW with only Thyroid with the best result.

In order to give a more thorough view on the ranking of the algorithm of Table \ref{result-comp}, Table \ref{rankings} is formed. In this table the best result on a dataset gets 1 and the worst gets a 5. In case of ties the algorithms get the same value and when computing the average rankings, the ties effect their averages as the mean of their respective ranks (if 2 algorithms are both ranked 3, they sum up as 3.5 to compute the average rank).
 
As can be seen in Table \ref{rankings}, CULM has the best rank of 1.9 which is far better than the second ranked Classical algorithms (rank 2.675). The third rank belongs to CULP with 2.925 and after that comes PgRkNN and HLCRW with 3.55 and 3.95 respectively. These are evidence that CULP and CULM are highly accurate classifiers and competitive with classical and similar works.

\begin{table}[ht]
\caption{Rankings of the algorithms of Table \ref{result-comp}.}\label{rankings}
\centering
\begin{tabular}{lccccc}
\toprule
\textbf{Dataset} & \textbf{CULP} & \textbf{CULM}& \textbf{Classical} & \textbf{HLCRW}&\textbf{PgRkNN}\\
\midrule
Zoo				&5                  &2	           &2      &   4   &    1      \\
Hayes       	&3                   &2          &1     &  5 &   4 \\
Iris 			&1               &2            &4       & 3 &  5  \\
Teaching        &4                  &1     &3    &  2 &  5 \\
Wine			&3                 &2    &1     &  5  &   4  \\
Sonar       	&2                   &1      &3     &  5 &    4  \\
Image       	&3                   &1       &2     &   5&   4  \\
Glass       	&2                   &1      &4    & 3&   5   \\
Thyroid     	&3            &3      &5     &1&    2  \\
Ecoli       	&2                   &1      &3     &   4 &    5  \\
Libras      	&5                   &4      &2      &  3 &  1 \\
Balance     	&3                  &1      &5     &  2&   4 \\
Pima        	&3                   &2       &1      &  4 &   5 \\
Vehicle     	&3			 	&2      &1     &  5&   4   \\
Vowel           &4                   &2      &1     &   5 &  3   \\
Yeast       	&1					&2     &3     &   4  &   5 \\
RedWine     &5 		        &3      &2     &  4 &   1\\
Segment         &2         &1         &3     & 5 &  4\\
Optical      &2                &3         &4     &   5     &   1  \\
Poker      &2		        &1     &3     & 4       &5          \\
\midrule
Average& 2.925& 1.9&   2.675& 3.9&  3.6\\
\bottomrule      
\end{tabular}
\end{table}

\section{Conclusion}
In this work, we proposed a novel way to look at the problem of classification using a link prediction scope. Our proposed memory efficient graph data structure LEG enabled the use of any link predictor to assist the classification procedure and captured not only the unlabeled and labeled data, but also the classes in a unified manner.

Our proposed algorithm CULP can be used with any link predictor to derive the class of the unlabeled data. In this work Common Neighbors, Adamic-Adar Index and resource allocation were used along with our own local link predictor called Compatibility Score as the predictors for CULP. Our algorithm demonstrated superiority to similar algorithms which use graph representations to classify a data point and our Compatibility Score was also one of the best predictors in our experiments.

We also extend CULP by a weighted majority vote with weights proportional to the probabilities of the predictions. CULM is the name of our extension which not only uses multiple predictors but it also exploits the low level features of the data as well.

Our experiments on both CULM and CULP showed high accuracy on 20 different datasets and superiority on all the classical approaches and similar graph based methods.

\section{Future Works}
There are a lot to be done with all the proposed methods and algorithms elaborated in this paper. We are going to test our Compatibility Score on graph datasets and test its accuracy on explicit link prediction problems. Another idea in our agenda is testing both CULP and CULM algorithms with other link prediction methods, possibly more complex ones such as random walk or matrix factorization to analyze any further improvement. Finally, a stacking approach to find the weights of CULM is under construction which hopefully be discussed in another work.

\bibliography{./References}
\bibliographystyle{ieeetr}

\end {document}